\documentclass[10pt,twocolumn,letterpaper]{article}

\usepackage[pagenumbers]{wacv} 

\usepackage{graphicx}
\usepackage{amsmath}
\usepackage{amssymb}
\usepackage{booktabs}

\usepackage{bm}
\usepackage{csquotes}
\usepackage{relsize}
\usepackage{xcolor}

\usepackage{tabularray}
\UseTblrLibrary{siunitx}

\SetBlockEnvironment{smallquote}

\newcommand\novVLM{{\mathit{nov}_{\mathrm{VLM}}}}
\newcommand\novPar{{\mathit{nov}_{\mathrm{p}}}}
\newcommand\accAbs{{\mathit{acc}_{\mathrm{abs}}}}
\newcommand\accRel{{\mathit{acc}_{\mathrm{rel}}}}

\newenvironment{smallitemize}{\vspace*{-4pt}
   \begin{itemize}
   \setlength{\parsep}{0pt}
   \setlength{\parskip}{0pt}
   \setlength{\topsep}{0pt}
   \setlength{\itemsep}{2pt}}
  {\end{itemize}

   \vspace*{-4pt}}

\newenvironment{smallenumerate}{\vspace*{-4pt}
   \begin{enumerate}
   \setlength{\parsep}{0pt}
   \setlength{\parskip}{0pt}
   \setlength{\topsep}{0pt}
   \setlength{\itemsep}{2pt}}
  {\end{enumerate}

   \vspace*{-4pt}}

\usepackage[pagebackref,breaklinks,colorlinks]{hyperref}

\usepackage[capitalize]{cleveref}
\crefname{section}{Sec.}{Secs.}
\Crefname{section}{Section}{Sections}
\Crefname{table}{Table}{Tables}
\crefname{table}{Tab.}{Tabs.}

\def\wacvPaperID{444} 
\def\confName{WACV}
\def\confYear{2026}

\begin{document}

\title{Estimating Image Creativity}

\author{Nisan Chhetri\\
NC State University\\
Raleigh, NC, USA\\
{\tt\small nchhetr@ncsu.edu}
\and
Christopher G. Healey\\
NC State University\\
Raleigh, NC, USA\\
{\tt\small healey@ncsu.edu}
\and
Sina Baghbanijam\\
NC State University\\
Raleigh, NC, USA\\
{\tt\small sbaghba@ncsu.edu}
}
\maketitle

\begin{abstract}
\noindent
What makes an image creative? Although this may seem like a highly subjective decision, research in psychology suggests that creativity may be based on a set of fundamental ``creativity dimensions'' such as novelty, surprise, and arousal. A deeper understanding of image creativity can benefit both research and practical applications. Exploring how human viewers use creativity dimensions to evaluate image creativity may allow us to predict a target image's human-rated creativity automatically. We present a novel framework to estimate the perceived creativity of an image by: (1) predicting the strengths of an image's novelty, surprise, and arousal creativity dimensions, and (2) combining the creativity dimension strengths to estimate an image creativity score. Controlled human-subject experiments were conducted on Amazon Mechanical Turk to collect data to train supervised machine learning models to estimate an image’s strengths on the three creativity dimensions. A final machine learning model takes the predicted dimension strengths as input to produce an estimated image creativity score. Results showed that our estimated creativity scores are statistically equivalent to human ratings of creativity for the images we tested. The novel contributions of our research are: (1) identification and validation of three psychological creativity dimensions used to estimate image creativity; (2) supervised machine learning models to automatically estimate novelty, surprise, and arousal strengths for an input image; (3) a supervised machine learning model to predict a creativity score from novelty, surprise, and arousal strengths for an input image, and (4) experimental methods to confirm the accuracy, advantages, and limitations of our approach.
\end{abstract}

\section{Introduction}
\label{sec:intro}

\noindent
This paper describes an investigation of image creativity. We are interested in estimating the creativity a viewer associates with an image. The Oxford English Dictionary defines creativity as ``$\ldots\,$the use of imagination or original ideas, especially in the production of an artistic work.'' Creativity is a subjective judgement, but across a group of viewers, it may be possible to determine an accurate aggregate estimate. For example, in previous work, we showed how to systematically increase perceived image aesthetic by varying the complexity of simulated paint strokes in a nonphotorealistic visualization, improving long-term memory for important data patterns \cite{Hea_TOG:2004,Tat_NPAR:2007}.
A better understanding of image creativity offers important advantages in both the research and the practical domains. It will confirm hypotheses about creativity and provide insights into how viewers perceive creativity at an image property level.

Creativity has been studied extensively in psychology, leading to the proposal of creativity dimensions: basic properties that form a foundation for evaluating creativity. Examples include novelty, arousal, surprise, and value \cite{boden1998creativity,franceschelli2022deepcreativity,horn2006product,gu2018neural,becattini2017surprise}. If true, a small set of dimensions may exist that represent image creativity. Estimating the perceived strengths of these dimensions in an image may allow us to combine them to predict an image creativity score.

\begin{figure*}[!t]
\includegraphics[width=1\textwidth]{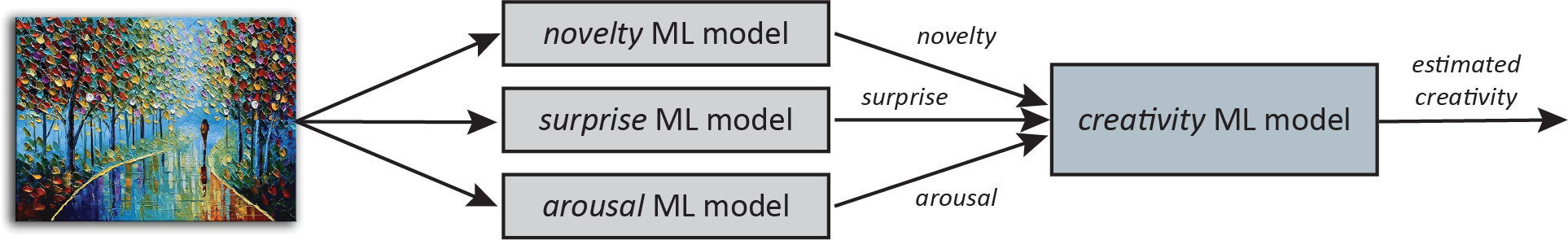}
\caption{Proposed image to creativity score pipeline: ML models estimate an image’s creativity dimension strengths, which are then combined in a final ML model to return an image creativity score}
\label{pipeline}
\end{figure*}

We propose a novel framework to automatically estimate the creativity of an image using creativity dimensions defined in the literature (Fig.~\ref{pipeline}). We first select creativity dimensions that we believe are most relevant to images. We modify the dimensions' definitions to fit the context of an image, then conduct a series of controlled human subject experiments to: (1) confirm correlations between an image’s dimension strengths and its creativity score; and (2) identify methods to estimate dimension strengths for a target image. We used Amazon Mechanical Turk (MTurk) to recruit participants to rate the strength and creativity of a curated collection of images for each of our creativity dimensions. Statistical analysis confirmed significant relationships between the strengths of the creativity dimensions and the creativity score. We used our experimental data to construct supervised machine learning (ML) models to estimate each dimension’s strength for a target image. Once the models for estimating dimension strengths were built, we designed a final ML model to predict an image's creativity score based on its dimension strengths. The result is a pipeline that automatically estimates an image's creativity using psychological creativity dimensions. The major steps of our proposed research are:

\begin{smallenumerate}
\item Understanding and identifying basic psychological dimensions of image creativity.
\item Conducting controlled MTurk experiments to obtain creativity dimension strengths and corresponding creativity scores for a set of curated images for each creativity dimension.
\item Constructing and validating predictive ML models for each creativity dimension using MTurk data.
\item Designing an ML model that uses estimated creativity dimension strengths to predict a creativity score for a target image.
\end{smallenumerate}

\noindent
We estimate creativity dimension strengths and then a creativity score for an image, rather than estimating creativity directly, for three reasons. First, this allows us to validate that basic creativity dimensions control the creativity of a target image. Second, maintaining independent creativity dimensions allows us to construct custom ML models for each dimension, rather than designing a single model that must simultaneously consider all three dimensions to estimate an image's creativity. This will be simpler to control and validate. Finally, our design allows us to explicitly specify creativity dimensions, rather than asking an ML model to identify them through feature extraction. The novel contributions of our work are as follows.

\begin{smallenumerate}
\item Identification and validation of creativity dimensions that can be used to estimate image creativity.
\item ML models to automatically estimate novelty, surprise, and arousal strengths from an input image.
\item An ML model to predict creativity from estimated novelty, arousal, and surprise strengths for an input image.
\item Experimental results that confirm the accuracy and highlight the advantages of our approach.
\end{smallenumerate}

\noindent
Our results offer important insights into creativity in general, and image creativity in particular. Creativity dimension strengths can be estimated and used to predict the creativity score a viewer associates with an image. Our proposed approach of vision language model (VLM)-based predictions and follow-on ML estimates may apply to additional creativity dimensions, potentially increasing the accuracy of our final creativity scores. The methods we use to decompose a complex problem like image creativity into fundamental steps, based on interdisciplinary research, provide a template for investigating similar problems.

The remainder of this paper identifies related work on creativity in psychology and on image creativity studies. We describe the design of MTurk experiments used to capture viewer-reported creativity dimension strengths and corresponding creativity scores for collections of 45 images per dimension. A comparison of convolutional neural networks (CNNs), mean shift, and VLM ML models found that a VLM was most accurate at predicting viewer-reported dimension strengths for a target image.
The estimated strengths act as input to a final ML model that estimates an image creativity score. Test sets held out from the MTurk experiments are used to validate both the dimension strength and the final creativity score models. Analysis shows our estimates are statistically comparable to human-reported creativity scores. Ablation experiments identify novelty as the main driver of image creativity scores. Based on these results, we suggest areas for potential future work to improve and extend our findings.

\section{Related Work}
\label{sec:related}

\noindent
 The term ``creativity'' has a rich history, being mentioned in many significant discoveries, music compositions, art, and other unique moments \cite{mehta2019creativity, sawyer2024explaining, carafoli2017complex, jones2012aha, nagy2020music}. There are two major creativity timelines: before and after 1950. Creativity originated from the Latin word ``creare'' with two interpretations \cite{still2016history}: a definition of natural forces and an association with God.

In 1950, J. P. Guilford presented a comprehensive model of creativity, arguing for its association with human traits \cite{guilford1950creativity}. This marked the beginning of the modern era of creativity. Although the term ``creativity'' is used in different ways over time, there is one similarity: a desire to explain a unique or original creation.
Researchers in the field of psychology define creativity as follows:

\begin{smallitemize}
\item ``A product or artifact generated by the person is creative if it is original and valuable for the person or the targeted audience \cite{guilford1967creativity}.''

\item ``Creativity is an ability to emerge with ideas or artifacts that are new, surprising, and valuable \cite{boden2005creativity}.''

\item ``Creativity is a psychological dimension normally distributed \cite{barbot2012creative}.''

\end{smallitemize}

\noindent
\cite{boden1998creativity} divides creativity into two broad categories. 

\begin{smallitemize}

\item Historical Creativity: Inventing new ideas or artifacts that have not been proposed previously.

\item Psychological Creativity: Smaller aspects of creativity that others might have invented but that an individual achieves for the first time.

\end{smallitemize}

\noindent
\cite{rhodes1961analysis} defines the term creativity with a 4Ps model:

\begin{smallitemize}

\item Person: Do correlations between habits, physique, attitudes, or other behaviors make a person creative?

\item Process: Does an underlying phenomenon, such as motivation, learning, or perception, motivate individuals to enhance conventional answers in creative ways?

\item Press: What environment supports creativity?

\item Product: What is the outcome of the creative process of a creative individual through a creative environment?

\end{smallitemize}

\noindent
Process-wise, the word creativity can take three different forms \cite{boden2004creative}.

\begin{smallitemize}

\item Combinational Creativity: A combination of two or more distinct ideas.

\item Exploratory Creativity: An exploration of a conceptual space defined by a set of rules.

\item Transformational Creativity: Formation of a new conceptual space by transforming the rules of an existing space.

\end{smallitemize}

\noindent
These interconnected representations show there is no standard definition of creativity. We chose to use the dimensions of product creativity, since our work is focused on estimating the creativity of images and not on the process of generating them.

\subsection{Creativity Dimensions}
\label{sec:dimension}

\noindent
Researchers have proposed various dimensions in the context of product creativity. The literature identifying these dimensions includes novelty, arousal, elaboration, value, and surprise. We chose novelty, surprise, and arousal as definitions that apply to images \cite{horn2006consumer}. Novelty represents the originality of the work, surprise captures the violation of expectations,
and arousal represents emotional activity.

\section{Human Subjective Ratings}
\label{sec:human}

\noindent
Modeling creativity is subjective and context-dependent, shaped by personal interpretation, cultural influences, and emotional response. Because of this, subjective strengths and scores provided by human raters are critical for the data needed to understand, quantify, and model creativity.
In psychology, one of the most widely accepted assessments is the Consensual Assessment Technique \cite{amabile1982social}, in which a creative product is defined as whatever observers see as creative. An observer might be an expert in the domain or a layperson, depending on the context and research goals. Numerous studies support the empirical validation of this technique and provide a basis for persuasive arguments that use perceptual data to represent creative artifacts.

Subjective ratings serve as training and evaluation data for supervised machine learning models that estimate creativity and its dimensions based on image properties. These human ratings are gathered through crowd-sourcing platforms for rapid, scalable data collection from a large, diverse population of online workers. MTurk \cite{AmazonMTurk} has been widely used in both computer vision and natural language processing for tasks involving aesthetics, humor, emotion, and creativity-–-all domains of inquiry that automatic metrics fail to capture.


\section{Estimating Creativity Dimension Strength}
\label{sec:strength}

\noindent
We begin by building supervised ML models to estimate novelty, surprise, and arousal strengths for a target image. Various methods could be used to perform this task. We focus on three commonly used and image-focused techniques: convolutional neural networks (CNNs), mean-shift anomalous pattern detection, and VLM-prompted strength estimation. The different algorithms are trained and tested using our MTurk datasets, cross-validation, and two different accuracy metrics.

\subsection{Convolutional Neural Networks}
\label{subsec:cnn}

\noindent
CNNs are a type of deep neural network designed to process pixel data arranged in a grid-like topology.
Since our interest is in image creativity and its dimensions, CNNs are an obvious candidate for estimating creativity strengths and scores.%
Different CNN architectures were evaluated for predicting creativity dimension strengths and creativity scores.

\vspace*{4pt}
\noindent
\textbf{ResNet-18 and ResNet-50.}
\label{subsec:resnet}
ResNet-18 and ResNet-50 are CNN architectures from the \textit{Residual Network} family \cite{he2016deep} that address the vanishing gradient problem in deep networks.
ResNet-18 is less computationally intensive, striking a balance in performance versus speed for feature extraction and limited computational resources environments.
ResNet-50 is a deeper model with a greater capacity to learn complex patterns in image data.

An important question is a CNN’s ability to capture subjective or semantic image features.
Creative dimensions like novelty may need to consider how common an image is in the real world, or how ``photographic'' an image appears.
We investigate whether a CNN architecture can recognize subjective patterns rather than objective, pixel-based ones.

\subsection{Mean-Shift}
\label{subsec:mean-shift}

\noindent
The mean-shift algorithm \cite{hermann2022fast} is used to identify outliers.
A patch-based representation allows the model to detect variations in anomalies across local and global patterns.
Mean-shift is flexible and efficient. It can identify novel data both globally, where an anomaly is detected across an entire image, and locally, where defects or anomalies occur at defined locations.
It is a compelling solution at larger scales, as it does not involve costly modifications to the network structure.
Given the potential need for anomaly detection in creativity dimensions, mean-shift appears to be a strong contender for our image analysis pipeline.

\subsection{Large Language Models}
\label{subsec:llm}

\noindent
VLMs are neural networks based on LLMs originally used for text processing. More recent versions are multi-modal, supporting analysis of images and video streams. In contrast to earlier types of natural language processing approaches, which typically used grounded rules or limited, specialized data, VLMs are pre-trained on very large corpora and later fine-tuned for specific applications. Thus, they produce generation-typical functionality, such as text creation, summarization, translation, or question answering. VLMs can be scaled across different domains, allowing them to generalize and perform well across diverse tasks.

We are interested in VLMs as they can analyze subjective multi-modal prompts, for example, ``Please examine this image and rate it on a novelty scale from one to seven. Novelty of an image refers to the likelihood that the image would be seen in the real world.'' Our proposed approach uses a VLM, specifically GPT-o3, to predict the strength of a creativity dimension. The VLM takes an image for analysis and a prompt generated from participant responses describing how they rated each of the creativity dimensions. VLMs can analyze text descriptions alongside visual information from images, which may give them greater capacity to subjectively assess dimensions of creativity, compared with computational image analysis methods such as CNNs. Competing methods are usually limited to identifying patterns in the image data. They can struggle to recognize higher-level subjective properties such as ``unusual'' or ``something that is not common in the real world.''

\section{Creativity Dimension Strength Models}
\label{subsec:estimate}

\noindent
We began by conducting a model selection experiment with 11 participants to compare three models: CNN, mean-shift, and VLM. Participants provided a novelty strength and a corresponding creativity score 45 pre-selected images spanning a Likert scale of $1\mbox{--}7$ with a semi-uniform distribution. CNN, mean-shift, and VLM models were constructed from the $11 \times 45 = 495$ results, then evaluated using two accuracy metrics. An VLM with a prompt explaining how humans determined novelty strength was the only model to return accurate estimates. We describe the construction and evaluation of the CNN, mean-shift, and VLM models, and summarize building VLM models using MTurk experiments that investigated novelty, surprise, and arousal.

\subsection{Novelty}
\label{subsec:novelty}

\noindent
Recall that psychology defines novelty as how new or unique an image feels based on what we see in everyday life. Images were collected from Internet-based sites such as Unsplash and Wiki-media Commons, and through general searches. We created a candidate list of approximately 80 images and narrowed it down to 45 candidates. The complete list of images was reviewed to confirm that they were clearly identifiable, free of visible artifacts, and licensed for public use. Images were selected to span the full $1\mbox{--}7$ range of novelty strength prior to participant data collection. Several categories of images were excluded: those with low resolution or compression artifacts, images depicting specific cultural practices or nudity, graphic violence, or strong emotional content. To span the range of possible novelty strengths, we subjectively assigned 7, 7, 6, 5, 6, 7, and 7 of our 45 images to the strength categories $1\mbox{--}7$, respectively.

\vspace*{4pt}
\noindent
\textbf{Experiment Design.}
Our experiment was designed in Qualtrics. We configured the following settings to ensure data quality and consistency across participants.

\begin{smallitemize}
\item \textit{Forced response:} Participants must score every image.
\item \textit{Random order:} Image presentation order is controlled by a Qualtrics randomized block.
\item \textit{One response per IP:} No duplicate participants.
\item \textit{Mobile device block:} Minimum screen size is required.
\item \textit{Bot detection}.
\end{smallitemize}

\noindent
Participants were told the purpose of the study, how to enter their dimension strengths and creativity scores for each image, and given our definition of novelty. Once they acknowledged they understood the instructions, they began the experiment. Two trap questions were embedded to detect inattentive participants. Trap questions displayed an image with an explicit instruction to select a specific score. Participants who failed more than one trap question were excluded from the analysis.

\vspace*{4pt}
\noindent
\textbf{Post-Experiment Questionnaire.}
After participants rated the 45 images, they completed an 11-question survey designed to explore their underlying thought processes when rating the images. Questions about the clarity of the experiment, which image properties participants used to choose a novelty strength, and the relationship between novelty and creativity were included. The survey allowed us to identify potential problems in the experiment, but none were found. Answers from the survey were also used to design the prompt for VLM-based novelty estimation.

\vspace*{4pt}
\noindent
\textbf{Experiment Responses.}
We analyzed the data to assess the reliability of the novelty strengths and their relationship with the creativity scores. We first removed outliers using the IQR method, discarding 11 pairs and retaining $495 - 11 = 484$. We then calculated a Pearson correlation coefficient between the images' novelty strengths and creativity scores, producing a statistically significant $r=0.71, p < 0.0001$. Mean novelty strengths per image ranged from $1\mbox{--}5$. Although individual participants rated some images with novelty strengths of $6$ and $7$, no image's mean strength reached this level. This range compression likely reflects conservative rating behavior among participants who lacked calibration examples to anchor the upper end of the scale.

\vspace*{4pt}
\noindent
\textbf{CNN Model.}
We investigated both ResNet-18 and ResNet-50 for constructing our CNN model. Since CNN performance improves with increased training data, we augmented our results by performing forty $9^\circ$ rotations on each image, producing $40 \times 45 = 1800$ unique images, each with a (\textit{novelty strength},$\;$\textit{creativity score}) pair. Rotation 
preserves color patterns, texture patches, and luminance distributions. 
If certain edge orientations are more important or occur more frequently, they should still be detected in the rotated images using different kernel filters. Images were randomly divided into three groups to perform a $\frac{2}{3}\mbox{--}\frac{1}{3}$ train--test split with three-fold cross-validation, re-run 50 times to ensure the model was not biased by a particular set of training and test data.

We explored a wide range of parameter settings for both ResNet-18 and ResNet-50 . The performance of CNNs was poor, however, leading to overfitting and unstable training. Accuracy was consistently in the 25--65\% range, better than a chance rate of 14\%, but still well below what we considered acceptable for a predictive model even in the best case.

\vspace*{4pt}
\noindent
\textbf{Mean-Shift Model.}
Mean-shift returns anomaly scores rather than novelty strength predictions. Before we considered how to use mean-shift scores to predict novelty, we examined them to see if they were likely to produce accurate results. The mean-shift algorithm needs a baseline image collection to act as a ``representative'' of normal images. We used CIFAR-10 to serve this purpose. Next, we calculated the average, standard deviation, minimum, and maximum mean-shift anomaly scores for images in each novelty strength category. This returned average anomaly scores of $(0.64, 0.63, 0.75, 0.61, 0.51)$ and standard deviations of $(0.18, 0.14, 0.62, 0.19, 0.15)$ for novelty strengths of 1, 2, 3, 4, and 5, respectively. The minimum and maximum anomaly scores across all 45 images were 0.31 and 2.32, defining an anomaly range of $2.32 - 0.31 = 2.01$.

The standard deviations in anomaly scores across novelty strength categories were 8.9\%, 6.9\%, 30.8\%, 9.4\%, and 7.5\% of the maximum possible anomaly difference. The high standard deviations suggest that anomaly scores were not consistent within any of the novelty strength categories. Anomaly scores varied widely within every novelty strength category. Given this, we did not feel the anomaly scores would be good predictors of novelty strength, so we did not proceed further to them into predicted novelty strengths.

\vspace*{4pt}
\noindent
\textbf{VLM Model.}
The failure of both the CNN and mean-shift models highlighted a fundamental mismatch. Both approaches compute statistical patterns from pixel values, but novelty is a semantic judgment about how rare an image is relative to everyday experiences. Predicting subjective judgments requires real-world knowledge. We therefore investigated VLMs as a third potential approach. VLMs use an instruction prompt to describe a problem to address. We also include an image referenced by the prompt.

The advantage of a VLM is our ability to ask subjective questions about a target input image. We used participants' questionnaire responses to understand their thought processes for estimating novelty strength. This identified three key ideas that participants used to rate novelty.

\begin{smallenumerate}
\item Familiar versus unique.
\item Plain background versus real background.
\item ``Real'' versus ``artificial.''
\end{smallenumerate}

\noindent
We integrated these insights to construct a prompt that instructed the VLM to return a novelty strength on the range of $1\mbox{--}100$ for each image.

\begin{displayquote}
Hi, I want you to give me a score from 1 to 100 for the image I upload here. Try to use the full score range if applicable. You have to provide the score based on three criteria:
1. Familiarity vs Uniqueness: How familiar is the image to everyday life/how often will you see it? More familiar or more often gets lower scores. More unique or hard-to-see often will get more scores.
2. Has background or not: Does the image have a plain background? For example, completely white or black or of any color. If the image has a background such as plain white or black, then it will get a low score. If not, it will get a high score.
3. Real vs not real: Is image photographed or, computer generated or stylized image. If the image is photographed or normally captured, then it will get a low score; if not, it will get a high score.
Did you understand my instructions? If yes, then I will upload the image and you will give me the score one by one. Ensure that your score isn’t influenced by the order of the images I upload.
\end{displayquote}

\noindent
The $1\mbox{--}100$ range was provided to allow the VLM finer-grained differentiation of novelty strength. Interestingly, the VLM never returned strengths below 5 or above 95. To account for this and for the average participant response range of 1–5, the VLM's strength estimates $\novVLM$ were scaled to a range that matched the participants’ results.

\vspace*{-4pt}
\begin{equation}
\novVLM = \lfloor ^{(\novVLM - 5)} / _{(95 - 5) \times 4} \rceil + 1
\end{equation}
\vspace*{-4pt}

\noindent
This maps $\novVLM$ to a range $[1,5]$, identical to the participants' average strength range. Two accuracy measures were then used to evaluate $\novVLM$ versus participants' average strengths $\novPar$.

This returns a transformed VLM strength on the range of 1 to 5, identical to the participants’ average strength range. We then calculated accuracy for each image's $(\novVLM, \novPar)$ pair in two ways.

\vspace*{-4pt}
\begin{align}
\accAbs &= \begin{cases} 1 \textrm{ if } \novVLM = \novPar \\ 0 \textrm{ otherwise} \end{cases}\\
\accRel &= 1 - \left[ ^{(\novVLM - \novPar)} / _{\max(\novPar - 1, 5 - \novPar)} \right] \label{acc_rel}
\end{align}
\vspace*{-4pt}

\noindent
$\accAbs$ calculates a binary accuracy: $1$ if $\novVLM$ matches $\novPar$, $0$ otherwise. $\accRel$ weights accuracy based on the maximum possible difference between $\novVLM$ and $\novPar$. For example, suppose the average participant novelty for an image is $\novPar=2$. Since the VLM returns estimates on the range $[1,5]$, the denominator for $\accRel$ is $\max(\novPar-1,5-\novPar)=\max(2-1,5-2)=3$, because the largest estimation error the VLM can make is $3$. $\accRel$ increases as $\novVLM \to \novPar$. When $\novVLM = \novPar$, $\accRel=1$, identical to $\accAbs$.

For VLM estimates, $\accAbs=49\%$ and $\accRel=77\%$, much higher than for either the CNN or mean-shift models, and above a 70\% threshold we considered acceptable for estimation accuracy. Given this, we used VLM-based models to estimate creativity dimension strengths moving forward.

\section{Creativity Estimation}
\label{sec:creåtivity}

\noindent
We conducted a second experiment on MTurk to collect (\textit{dimension strength},$\;$\textit{creativity score}) pairs for the three creativity dimensions novelty, surprise, and arousal.

\subsection{Dimension Definitions and Image Datasets}
\label{subsec:defn}

\noindent
Separate definitions and sets of 45 images were used for each dimension. For novelty, the same definition and 45 images from the pre-experiment were used. Interestingly, the post-experiment questionnaire did not identify plain versus real background as a key idea used to rate novelty. It was therefore removed from the prompt asking a VLM to rate the novelty of a target image.

Surprise was defined as:

\begin{displayquote}
Surprise measures how unexpected or implausible an image feels, based on your normal expectations. Rate an image higher if it has unexpected or implausible elements compared to what you typically expect. Rate an image lower if it is predictable or expected.
\end{displayquote}

\noindent
Images were collected using the Google Veo image generation model to span the range of surprise categories, based on our subjective judgment. Low-surprise images included plausible scenes such as a plate lying on a table with a spoon. High-surprise images included impossible and highly unexpected scenes such as a frog playing guitar.

Because surprise is grounded in implausibility and expectation violation, AI-generated images allow the creation of scenes ranging from completely ordinary to physically impossible, something difficult to achieve with photographs.

Arousal was defined as:

\begin{displayquote}
Arousal refers to the intensity of the emotional response an image evokes, regardless of whether the feeling is positive or negative. Rate arousal higher if the image triggers a strong, energized emotional reaction (e.g., feeling alert, activated, or charged). Rate arousal lower if the image evokes a weak, minimal emotional reaction (e.g., feeling still, calm, or peaceful). When making your ratings, you should ignore whether the picture is positive or negative, good or bad. Instead, we ask you to rate the picture only in terms of the emotional intensity that it evokes. As an example, imagine two pictures: (1) an athlete winning a gold medal (something good); (2)  an athlete getting injured and losing a race (something bad). You may give both images the same rating if they create a similar arousal or emotional response, even though the feelings of pleasure are different. Your answers should be based only on the emotional intensity the picture invokes.
\end{displayquote}

\noindent
We used the Open Affective Standardized Image Set (OASIS) \cite{kurdi:2017}, an image collection rated on valence and arousal, as the source for our 45 arousal images.
We uniformly sampled the arousal dimension to select the 45 images used in our experiment\footnote{The code, image datasets, and (\textit{dimension strength}, \textit{creativity score}) pairs we used will be made available through a public GitHub repository.}.

\subsection{MTurk Experiment}
\label{subsec:mturk_exp}

\noindent
We recruited US-based Master Turkers\footnote{98\%  lifetime HIT rate approval.}, who have provided high-quality results during past investigations. Qualtrics bot detection, IP address duplication, and device restrictions were applied as before. A compensation rate of \$10 was provided for completing the experiment, which took between 25 and 35 minutes. A total of 60 Master Turkers (20 per creativity dimension) responded over a four-day period.

The three experiments followed the same model-selection structure: instructions, dimension definition, rating with trap questions, and a follow-up questionnaire. Participants’ questionnaire responses served as the primary source of explanations for each dimension, used in our VLM prompt during dimension strength estimation.

\subsection{MTurk Results}
\label{subsec:mturk_res}

\noindent
IQR results identified 31, 27, and 35 outliers in the 900 novelty, surprise, and arousal (\textit{strength},$\;$\textit{creativity score}) pairs. Pearson correlations of $r=0.82, p < 0.0001$, $r=0.76, p < 0.0001$, and $r=0.60, p < 0.0001$ were obtained for novelty, surprise, and arousal strengths versus creativity scores, confirming positive relationships in all three instances. The range of average dimension strengths for each image was $[1,6]$, $[1,6]$, and $[1,7]$ for novelty, surprise, and arousal, respectively. These were used to calculate $\accRel$ (Eq.~\ref{acc_rel}) for each dimension.

\subsection{VLM Estimates}
\label{subsec:llm_estimate}

\noindent
Three VLM models were constructed using the MTurk experiment results to estimate novelty, surprise, and arousal strengths. The main differences during model construction were the image dataset used and the prompt given to the VLM. For novelty, the image dataset and prompt were identical to those used during model selection. For surprise, participants' questionnaire responses identified: (1) plausibility versus impossibility; (2) ordinary versus out-of-place, and (3) seen versus unseen as common reasoning patterns when choosing surprise strengths. This produced the following surprise prompt.

\begin{displayquote}
Hi, I want you to give me a score from 1 to 100 for the image I upload here. Try to use the full score range if applicable. You have to provide a combined score from 1 to 100 based on three criteria: plausibility versus impossibility: Does the image depict a scene that could happen within the laws of our reality? Realistic and physically possible scenes receive lower surprise ratings; scenes that are absurd, surreal, or impossible receive higher ratings; ordinary versus out-of-place: Does the image depict a subject in its expected context and behaving predictably? Ordinary scenes with subjects acting as one would normally expect receive lower surprise ratings; scenes where the subject is out of place or acting uncharacteristically receive higher ratings; seen versus unseen: Has the viewer been exposed to this concept or a similar idea before? Concepts the viewer has seen before or that feel common receive lower surprise ratings; concepts that feel entirely new, unique, or personally unseen receive higher ratings. Did you understand my instructions? If yes, then I will upload the image and you will give me the score one by one. Ensure that your score isn’t influenced by the order of the images I upload.
\end{displayquote}

\noindent
For arousal, participants' questionnaire responses identified: (1) emotional versus neutral; (2) compelling versus mundane, and (3) impactful versus forgettable as common reasoning patterns when choosing arousal strengths. This produced the following arousal prompt.

\begin{displayquote}
Hi, I want you to give me a score from 1 to 100 for the image I upload here. Try to use the full score range if applicable. You have to provide a combined score from 1 to 100 based on three criteria: emotional versus neutral: How intensely does the image stir up feelings? Images that generate strong emotions like awe, fear, or joy receive higher arousal ratings; those that feel emotionally neutral receive lower ratings; compelling versus mundane: Does the image depict an engaging subject or a simple object? Images with dynamic scenes or striking subjects that tell a story receive higher arousal ratings; images of everyday, mundane objects receive lower ratings; impactful versus forgettable: Does the image trigger an immediate gut reaction? Images that grab your attention instantly and leave a lasting impression receive higher arousal ratings; those that are forgettable or are met with indifference receive lower ratings. Did you understand my instructions? If yes, then I will upload the image and you will give me the score one by one. Ensure that your score isn’t influenced by the order of the images I upload.
\end{displayquote}

\noindent
\textbf{Accuracy.}
The novelty, surprise, and arousal VLM models achieved accuracy scores of $\accAbs=(67\%, 13\%, 4\%)$ and $\accRel=(90\%, 76\%, 72\%)$, respectively.

\section{Estimating Creativity}
\label{sec:creativity}

\noindent
The final step is to convert novelty, surprise, and arousal strengths for an image into a creativity score. We constructed supervised ML models that take the three dimension strengths as input and return a creativity score estimate on the range $[0,7]$ as output. We used our VLM models to estimate novelty, surprise, and arousal strengths for the $45 \times 3 = 135$ images from the MTurk experiments. This allowed us to calculate: (1) creativity score versions of $\accRel$ for an image's estimated score versus its participant-reported score using VLM-estimated or participant-reported dimension strengths\footnote{Participant-reported strengths are supplemented with VLM-predicted strengths, \textit{e.g.,} for a novelty image the novelty strength is participant-reported but the surprise and arousal strengths are VLM-predicted, since these were not provided by participants during the novelty experiment.} as input, and (2) $t$ and $p$-values comparing the model-estimated scores for the two input types. Four supervised ML approaches were considered: random forest (RF), support vector classification (SVC), $k$-nearest neighbors (KNN), and a single layer artificial neural network (ANN). The same 50 randomized repetitions of three-fold cross-validation were used during model training.

\begin{table}[!t]
\caption{$\accRel$ for RF, SVC, KNN, and ANN creativity score predictions using VLM-estimated strengths or participant-report strengths as input; $t$ and $p$-values for VLM-based versus participant-based creativity scores.}
\label{table:three_dim}
\centering
\begin{tblr}{
  colspec={c|c c Q[si={table-format=+-1.2},c] c},
  rowspec={Q[m]Q[m]Q[m]Q[m]Q[m]},
  row{odd}={gray9}, row{even}={blue9},
  row{1}={gray7},
  columns={colsep=5pt}
}
\textbf{Model} & \textbf{VLM} & \textbf{Human} & $t$ & $p$\\
RF &
77.8\% & 79.7\% & 0.06 & 0.95\\
SVC &
77.2\% & 79.9\% & 1.31 & 0.19\\
KNN &
77.6\% & 80.7\% & -1.34 & 0.18\\
ANN &
78.4\% & 79.2\% & 0.38 & 0.71\\
\end{tblr}
\end{table}

Each ML model used VLM-estimated or participant-reported dimension strengths to predict a creativity score for each image. Comparing the predictions to the score participants assigned to an image allowed us to calculate $\accRel$.

We also compared scores for VLM-predicted versus participant-reported dimension strength inputs. Table~\ref{table:three_dim} shows results for each of our four models. Two insights are immediately obvious: (1) $\accRel$ is stable across the four models, and (2) in no case are the models' creativity score predictions statistically significantly different across the two types of dimension strength inputs. This means that our creativity score models produce estimates that are statistically identical for VLM-estimated strength inputs versus human-reported strength inputs.

\subsection{Ablation Studies}
\label{subsec:ablatoin}

\noindent
To investigate the importance of each creativity dimension, we used the same model construction approach for RF, SVC, KNN, and ANN. We used as input estimated novelty, surprise, or arousal from our VLM strength models rather than combining all three strengths. Because we also collected participant-reported strengths for each dimension's 45 images, we compared $\accRel$\footnote{We did not consider $\accAbs$ in the ablation studies since $\accRel$ has been shown to be the more appropriate accuracy metric.} for creativity across three categories: (1) a single VLM-based strength; (2) a single human-reported strength; a combination of all three strengths (\textit{i.e.,} our initial creativity estimation results). Analysis of variance (ANOVA) was used to identify statistically significant differences in accuracy. If found, Wilcoxon post-hoc tests were run to identify significantly different (\textit{input source}, \textit{model}\thinspace) pairs. We only report pairs where either the model or the input source is identical (\textit{i.e.,} either a significantly better input for a given model, or significantly better models for a given input).

\vspace*{4pt}
\noindent
\textbf{Novelty.}
\label{subsec:ablation_nov}
For novelty as input, $F(11,484)=8.11, p< 0.001$. Wilcoxon post-hoc tests showed human input outperformed both VLM and combined dimensions input for all four models.
Combined input outperformed VLM for the SVC model.

\vspace*{4pt}
\noindent
\textbf{Surprise.}
\label{subsec:ablation_sur}
For surprise as input, $F(11,484)=8.3, p< 0.001$. Wilcoxon post-hoc tests showed human input outperformed both VLM and combined dimensions input for all four models, except for (\textit{human}, \textit{KNN})--(\textit{combined}, \textit{KNN}), which was not significantly different.
Combined input outperformed VLM for the KNN model.

\vspace*{4pt}
\noindent
\textbf{Arousal.}
\label{subsec:ablation_aro}
For arousal as input, $F(11,484)=2.32, p< 0.009$. However, none of the accuracy differences across common-source or common-model pairs were significant, suggesting comparable performance regardless of the chosen input source or model.

Results from the ablation studies showed several interesting results: (1) human input outperforms VLM or combined input in almost all cases for novelty and surprise; (2) model choice was not significant in any situation, and (3) input source or model choice was not significant for arousal. We concluded that human input outperforms VLM-based estimates as input for predicting an image's creativity score. This is further reinforced by situations where human input outperformed combined input. Recall that for combined input, two of the three dimensions are VLM-estimated. A more comprehensive ablation study would collect participants' strength ratings for novelty, surprise, and arousal for all 135 images, allowing us to test for single-dimension differences when the input was entirely human-rated. This is left as an area for future work.

\section{Conclusions and Future Work}
\label{sec:conclusion}

\noindent
This paper presents a method for estimating image creativity based on the creativity dimensions novelty, surprise, and arousal. We constructed CNN, mean-shift, and VLM-based models using participant-reported novelty strengths and creativity scores collected with an experiment that asked participants to rate 45 images spanning the range of novelty from $1$ (low) to $7$ (high). Only the VLM model produced high-quality predictions. We next ran novelty, surprise, and arousal experiments on MTurk to collect data for constructing three VLM-based strength models, one for each dimension. Model accuracy confirmed each model was capable of predicting participant-report results.

A final ML model used novelty, surprise, and arousal estimates to predict an image's creativity. We considered four supervised ML approaches: RF, SVC, KNN, and ANN. All four models produced acceptable estimation accuracies, demonstrating our ability to estimate participant-reported image creativity.

Future work is investigating additional creativity dimensions as sources for improving image creativity predictions. Because of how we designed our prediction pipeline, it is easy to add dimensions: an additional MTurk experiment is run to collecct (\textit{dimension strength}, \text{creativity score}) pairs for the images being tested, a post-experiment questionnaire is conducted to determine which image properties participants are considering when they report the image's dimension strength, and a VLM-based strength prediction model is constructed to prediction an input image's dimension strength. The prediction is used as an additional input to the final creativity score estimation model.

Another study is whether our approach of: (1) predicting foundational properties of a problem domain, and (2) combining those properties to predict a final value for the domain extends to areas outside of image creativity. For example, suppose we wanted to determine whether different visualization designs made it harder or easier for a viewer to answer questions about the underlying data. We might first build models to determine what type of visualization to use (\textit{e.g.}, bar chart, scatterplot, line chart), which visual properties to map the data to (\textit{e.g.}, hue, luminance, size, position), then combine these into an input for a model that integrates visualization type and visual feature choice to recommend an effective visualization. A similar approach could apply to any problem domain that can be structured as a combination of foundational properties.

Finally, although we restricted our participant pool to US-based workers to identify English-fluency and common past experiences and expectations, creativity and arousal are subjective and culturally dependent. Follow-on experiments to determine if and how cultural difference affect image creativity dimension strengths and creativity score ratings would be valuable and informative.


{\small
\bibliographystyle{ieee_fullname}
\bibliography{healey,nchhetr-thesis}

@string{TOG     = "ACM Transactions on Graphics"}

@article{Hea_TOG:2004,
  author =      {Healey, C. G. and Enns, J. T. and Tateosian, L. G. and Remple, M.},
  title =       {Perceptually-Based Brush Strokes for Nonphotorealistic Visualization},
  journal =     TOG,
  volume =      {23},
  number =      {1},
  pages =       {64--96},
  month =       {},
  year =        {2004}
}

@inproceedings{Tat_NPAR:2007,
  author =	{Tateosian, L. G. and Healey, C. G. and Enns, J. T.},
  title =	{Engaging Viewers Through Nonphotorealistic Visualizations},
  booktitle =	{Proceedings 5th International Symposium on Non-Photorealistic Animation and Rendering (NPAR 2007)},
  pages =	{93--102},
  address =	{San Diego, California},
  organization=	{},
  year =	{2007}
}

@article{horn2006product,
  title={Product Creativity: Conceptual Model, Measurement and Characteristics},
  author={Horn, D. and Salvendy, G.},
  journal={Theoretical Issues in Ergonomics Science},
  volume={7},
  number={4},
  pages={395--412},
  year={2006},
  publisher={Taylor \& Francis}
}

@article{rhodes1961analysis,
  title={An Analysis of Creativity},
  author={Rhodes, M.},
  journal={The Phi Delta Kappan},
  volume={42},
  number={7},
  pages={305--310},
  year={1961},
  publisher={JSTOR}
}

@article{amabile1982social,
  title={Social Psychology of Creativity: A Consensual Assessment Technique},
  author={Amabile, T. M.},
  journal={Journal of Personality and Social Psychology},
  volume={43},
  number={5},
  pages={997},
  year={1982},
  publisher={American Psychological Association}
}

@article{horn2006consumer,
  title={Consumer-Based Assessment of Product Creativity: A Review and Reappraisal},
  author={Horn, D. and Salvendy, G.},
  journal={Human Factors and Ergonomics in Manufacturing \& Service Industries},
  volume={16},
  number={2},
  pages={155--175},
  year={2006},
  publisher={Wiley Online Library}
}

@article{mehta2019creativity,
  title={Creativity: Past, Present, and Future},
  journal={Consumer Psychology Review},
  volume={2},
  number={1},
  pages={30--49},
  author={Mehta, R. and Dahl, D. W.},
  year={2019}
}

@book{sawyer2024explaining,
  title={Explaining Creativity: The Science of Human Innovation},
  author={Sawyer, R. K. and Henriksen, D.},
  year={2024},
  publisher={Oxford University Press}
}

@article{carafoli2017complex,
  title={The Complex Structure of the Creativity Process},
  author={Carafoli, E.},
  journal={Rendiconti Lincei},
  volume={28},
  pages={449--462},
  year={2017},
  publisher={Springer}
}

@book{jones2012aha,
  title={The {Aha!} Moment: A Scientist's Take on Creativity},
  author={Jones, D.},
  year={2012},
  publisher={JHU Press}
}

@article{nagy2020music,
  title={Music and Embodied Creative Space},
  author={Nagy, Z. S.},
  journal={Journal of Genius and Eminence},
  volume={5},
  pages={64--74},
  year={2020}
}

@inproceedings{still2016history,
  title={A History of Creativity for Future {AI} Research},
  author={Still, A. and d'Inverno, M.},
  booktitle={Proceedings of the 7th Computational Creativity Conference (ICCC 2016)},
  year={2016},
  organization={Universit\'{e} Pierre et Marie Curie}
}

@article{guilford1950creativity,
  title={Creativity},
  author={Guilford, J. P.},
  journal={American Psychologist},
  volume={5},
  number={9},
  pages={444--454},
  year={1950},
  publisher={Washington, DC}
}

@article{guilford1967creativity,
  title={Creativity: Yesterday, Today and Tomorrow},
  author={Guilford, J. P.},
  journal={The Journal of Creative Behavior},
  volume={1},
  number={1},
  pages={3--14},
  year={1967},
  publisher={Wiley Online Library}
}

@incollection{boden2005creativity,
  title={What is Creativity?},
  author={Boden, M. A.},
  booktitle={Creativity in Human Evolution and Prehistory},
  pages={27--55},
  year={2005},
  publisher={Routledge}
}

@article{barbot2012creative,
  title={Creative Thinking in Music: Its Nature and Assessment Through Musical Exploratory Behaviors},
  author={Barbot, B. and Lubart, T.},
  journal={Psychology of Aesthetics, Creativity, and the Arts},
  volume={6},
  number={3},
  pages={231},
  year={2012},
  publisher={Educational Publishing Foundation}
}

@article{boden1998creativity,
  title={Creativity and Artificial Intelligence},
  author={Boden, M. A.},
  journal={Artificial Intelligence},
  volume={103},
  number={1-2},
  pages={347--356},
  year={1998},
  publisher={Elsevier}
}

@book{boden2004creative,
  title={The Creative Mind: Myths and Mechanisms},
  author={Boden, M. A.},
  year={2004},
  publisher={Routledge}
}

@article{hermann2022fast,
  title={Fast and Efficient Image Novelty Detection Based on MeanSshifts},
  author={Hermann, M. and Umlauf, G. and Goldl{\"u}cke, B. and Franz, M. O.},
  journal={Sensors},
  volume={22},
  number={19},
  pages={7674},
  year={2022},
  publisher={MDPI}
}

@inproceedings{he2016deep,
  title={Deep Residual Learning for Image Recognition},
  author={He, K. and Zhang, X. and Ren, S. and Sun, J.},
  booktitle={Proceedings of the IEEE Conference on Computer Vision and Pattern Recognition},
  pages={770--778},
  year={2016}
}

@article{franceschelli2022deepcreativity,
  title={{DeepCreativity}: Measuring Creativity With Deep Learning Techniques},
  author={Franceschelli, G. and Musolesi, M.},
  journal={Intelligenza Artificiale},
  volume={16},
  number={2},
  pages={151--163},
  year={2022},
  publisher={SAGE Publications Sage UK: London, England}
}

@article{gu2018neural,
  title={The Neural Mechanism Underlying Cognitive and Emotional Processes in Creativity},
  author={Gu, S. and Gao, M. and Yan, Y. and Wang, F. and Tang, Y.-Y. and Huang, J. H.},
  journal={Frontiers in Psychology},
  volume={9},
  pages={1924},
  year={2018},
  publisher={Frontiers Media SA}
}

@article{becattini2017surprise,
  title={Surprise and Design Creativity: Investigating the Drivers of Unexpectedness},
  author={Becattini, N. and Borgianni, Y. and Cascini, G. and Rotini, F.},
  journal={International Journal of Design Creativity and Innovation},
  volume={5},
  number={1-2},
  pages={29--47},
  year={2017},
  publisher={Taylor \& Francis}
}

@misc{AmazonMTurk,
  author       = {{Amazon.com, Inc.}},
  title        = {Amazon Mechanical Turk},
  year         = {2005},
  howpublished = {\url{https://www.mturk.com/}},
  note         = {Accessed: 8 April 2025}
}

@article{kurdi:2017,
  title={Introducing the Open Affective Standardized Image Set ({OASIS})},
  author={Kurdi, B. and Lozano, S. and Banaji, M. R.},
  journal={Behavior Research Methods},
  volume={49},
  number={2},
  pages={457--470},
  year={2017}
}
}

\end{document}